%% file: main_icassp_medva.tex
\documentclass{article}

\usepackage{spconf,amsmath,graphicx,hyperref}
\usepackage{booktabs}
\usepackage{multirow}
\usepackage{amsfonts}

\title{A Medical Multimodal Diagnostic Framework Integrating Vision-Language Models and Logic Tree Reasoning}
{
\name{\normalsize 
Zelin Zang$^{1,2}$, Wenyi Gu$^{1}$, 
      Siqi Ma$^{2}$, 
      Dan Yang$^{3}$, 
      Yue Shen$^{3}$,
      Zhu Zhang$^{4}$,
      Guohui Fan$^{4}$, 
      Wing-Kuen Ling$^{1}$,
      Fuji Yang$^{1}$
}}
\address{ 
    $^{1}$ Tsientang Institute of Advanced Study (TIAS), Hangzhou, China \\
    $^{2}$ Westlake University, Hangzhou, China 
    $^{3}$ Ant Group, Hangzhou, China \\
    $^{4}$ China-Japan Friendship Hospital, Beijing, China 
    \texttt{zangzelin@westlake.edu.cn}, 
}

\begin{document} 

%
\maketitle
\begin{abstract} 
With the rapid growth of large language models (LLMs) and vision–language models (VLMs) in medicine, simply integrating clinical text and medical imaging does not guarantee reliable reasoning. Existing multimodal models often produce hallucinations or inconsistent chains of thought, limiting clinical trust.  
We propose a diagnostic framework built upon LLaVA that combines vision–language alignment with logic-regularized reasoning. The system includes an input encoder for text and images, a projection module for cross-modal alignment, a reasoning controller that decomposes diagnostic tasks into steps, and a logic tree generator that assembles stepwise premises into verifiable conclusions.  
Evaluations on MedXpertQA and other benchmarks show that our method improves diagnostic accuracy and yields more interpretable reasoning traces on multimodal tasks, while remaining competitive on text-only settings. These results suggest a promising step toward trustworthy multimodal medical AI.
\end{abstract}
\begin{keywords}
Medical Multimodal Diagnosis; Vision-Language Model; Logic Tree Reasoning; Explainable Artificial Intelligence; 
\end{keywords}
\section{Introduction}
\input{sec_intro.tex}

\section{Related Work}
\input{sec_relatwork.tex}

\section{Method}
\input{sec_method.tex}

\section{Experiments}
\input{sec_exp.tex}

\section{Conclusion}
\input{sec_conclusion.tex}
\textbf{Compliance with Ethical Standards.}
The authors used LLM to assist with language editing and polishing.
All technical content, results, and conclusions are solely authored and verified by the authors.

\newpage

\label{sec:refs}
 
\bibliographystyle{IEEEbib}
\bibliography{refs}

\clearpage
\appendix

\end{document}

%% file: sec_intro.tex
\label{sec:intro}

Deep learning has greatly advanced medical AI. Large language models (LLMs) \cite{singhal2023medpalm} and vision–language models (VLMs) \cite{tiu2022chexzero} can now jointly process clinical text, history, and images, achieving promising results across various medical domains including disease classification \cite{sun2022artificial} and clinical prediction \cite{zhou2021eleven}. This often improves predictions — for example, in our tests, models combining CT findings with symptom descriptions narrowed differential diagnoses better than text-only systems.  
Yet simply adding modalities does not guarantee sound reasoning. Benchmarks such as MedXpertQA \cite{zuo2025medxpertqa}, VQA-RAD \cite{lau2018vqarad}, PathVQA \cite{he2020pathvqa}, and PubMedQA \cite{jin2019pubmedqa} show that multimodal models may contradict themselves or ignore key evidence. Such inconsistencies limit trust and slow adoption, since physicians must verify how conclusions are reached \cite{belisle2024llmcare}.

\begin{figure}[t]
\centering
\includegraphics[width=0.5\textwidth]{fig/overview_medva.pdf}
\caption{\textbf{Overview.} Left: baseline VLMs rely on implicit CoT, causing hallucinations, inconsistency, and weak grounding.  
Right: our framework integrates vision--language alignment and logic-regularized reasoning, yielding traceable logic trees and consistent diagnoses.}
\label{fig:overview}
\end{figure}

Despite progress, strong VLMs still give confident but wrong answers, or hallucinations \cite{ji2023survey}, especially when text and image evidence conflict. In pilot runs, some even contradicted radiology reports by over-relying on text — dangerous in clinical practice, where unsupported predictions can mislead physicians.

On closer inspection, these errors seemed to stem from implicit pattern matching rather than structured reasoning, which made their outputs hard to audit. To address this, we tried making the reasoning process explicit instead of directly predicting a label. Early attempts were unstable, but adding formal logic constraints stabilized training and surprisingly improved both accuracy and interpretability. Our current model now learns to arrange premises step by step before reaching a conclusion — similar to how clinicians justify decisions in tumor boards — producing reasoning chains that are easier for physicians to inspect and question.

Building on LLaVA \cite{liu2023llava}, which has demonstrated strong vision-language alignment capabilities and efficient multimodal fusion, we designed a system with four cooperating parts: an input encoder for clinical text and CT/MRI images, a vision-language alignment module that maps image features into the language space, a reasoning controller that breaks down diagnostic tasks into intermediate steps, and a logic tree generator that assembles those steps into a verifiable premise-conclusion chain.
In early trials, we noticed that logic rules alone had limited impact — the model still ignored subtle visual cues. Multimodal alignment turned out to be essential: once we projected visual features into the same space as text, the reasoning controller produced much more consistent outputs. Our final design combines formal logic constraints with vision-language alignment, yielding reasoning trees that physicians can check. Experiments on MedXpertQA \cite{zuo2025medxpertqa} and other benchmarks confirmed the benefit: accuracy improved, and the reasoning traces were easier to verify, which we see as a step toward more trustworthy medical AI.

%% file: sec_relatwork.tex
\textbf{Vision–Language Models in Healthcare.} Recent multimodal studies have produced several specialized VLMs for medicine. CheXzero showed that paired image-text pretraining could reach radiologist-level performance on chest X-rays without manual labels, hinting that large unannotated corpora might replace expensive annotation. Med-Flamingo \cite{moor2023medflamingo} adapted OpenFlamingo to the clinical setting and reported a 20 \% improvement in blinded physician ratings when rationales were provided — a result that convinced us that explanation quality really matters in practice. LLaVA-Med \cite{li2023llavamed} further demonstrated that instruction-tuning on PubMed images with GPT-4 captions yields strong biomedical VQA performance. More recent systems such as UMIT \cite{yu2025umit} and HealthGPT \cite{lin2025healthgpt} attempt to unify multiple imaging tasks and support both comprehension and generation. Beyond diagnostic imaging, recent work has explored multimodal integration in other biomedical contexts such as spatial transcriptomics \cite{zang2025must} and single-cell analysis \cite{xu2025complex}, demonstrating the broader applicability of vision-language methods. These efforts show impressive gains in recognition, but we found that most still lack structured reasoning that clinicians can audit. 
\textbf{Logic-Based Reasoning in Medical LLMs.} Explicit reasoning has been explored to increase trust in model outputs. Med-PaLM \cite{singhal2023medpalm} and Med-PaLM 2 \cite{singhal2025medpalm2} fine-tuned general LLMs to produce step-by-step chains of thought, though these often remain opaque and difficult to verify. MedLA \cite{ma2024medla} goes further by using multi-agent dialogue to refine logic trees, while MDAgents \cite{kim2024mdagents} coordinates multiple expert agents under a moderator. These approaches improved clinical rigor but came with heavy computational overhead. We chose a simpler path: embedding logic constraints directly into a single-model chain-of-thought, aiming to keep the reasoning consistent and interpretable without requiring multi-agent orchestration.

%% file: sec_method.tex
\begin{figure*}
    \centering
    \includegraphics[width=0.99\linewidth]{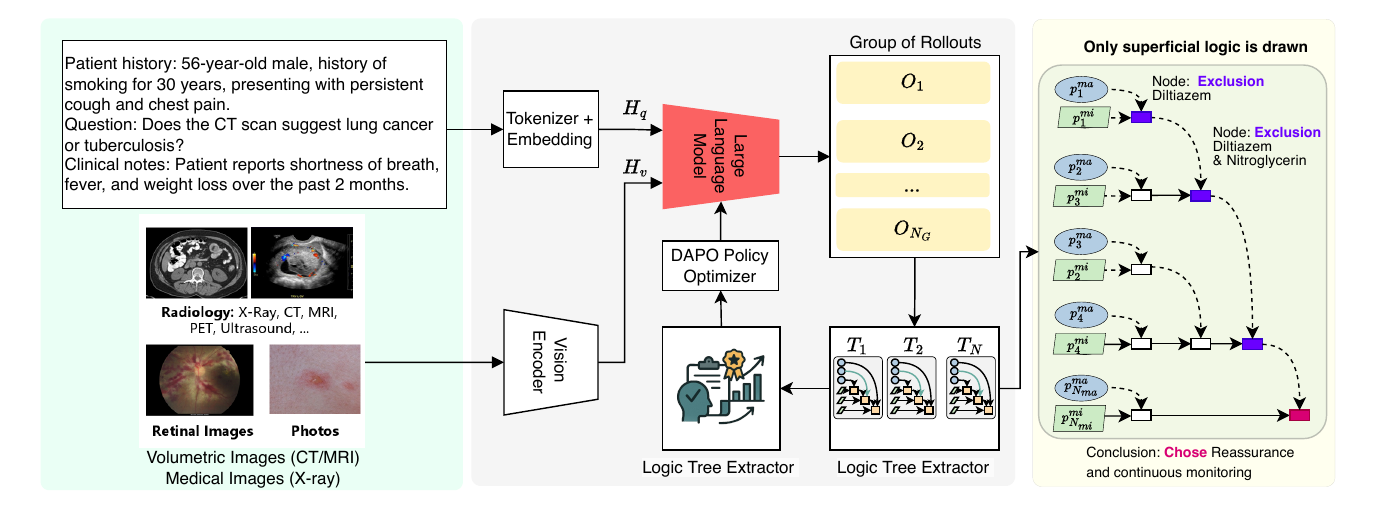} 
\vspace{-1.0em}
\caption{\textbf{Multimodal logic-regularized reasoning framework.} 
Patient history, clinical notes, and medical images (e.g., CT, MRI, X-ray) are encoded by text and vision encoders. 
Their embeddings are fed into a large language model with a DAPO policy optimizer, producing multiple candidate reasoning rollouts. 
Each rollout is parsed by the logic tree extractor into syllogistic premises and conclusions, forming verifiable logic trees. 
The final diagnosis is obtained together with a traceable reasoning chain, improving both accuracy and interpretability.}
\vspace{-1em}
    \label{fig:medva_arch}
\end{figure*}

\textbf{Overall Framework.}  
We extend LLaVA to the clinical setting by adding explicit logic regularization. Our system proceeds in three stages, illustrated in Figure \ref{fig:medva_arch}. First, we embed medical images and clinical narratives into a shared representation space so that visual and textual cues can interact early. Next, the model generates several candidate chains-of-thought, which we refine with a dynamic optimization scheme (DAPO) \cite{yu2025dapoopensourcellmreinforcement} that balances three signals: diagnostic accuracy, logical consistency, and image-text grounding. Finally, we parse the best reasoning path into syllogistic triads and assemble them into a logic tree that clinicians can inspect step by step.
During development, we tried freezing the vision encoder to save compute, but convergence became unstable and the model ignored subtle imaging findings. Jointly training all components solved this problem, so we adopted a fully end-to-end optimization strategy.

\textbf{Visual \& Text Encoder.}  
We adopt a ViT \cite{dosovitskiy2020vit} backbone pretrained on vision–language tasks as the image encoder. Each 2D medical image is split into patches and converted into visual tokens ${v_1,\dots,v_M}$. For volumetric scans (CT or MRI), we process each slice with the same encoder and then use a slice-fusion transformer with multi-head attention to combine information across slices. In early tests, we tried simply averaging slice features, but this blurred subtle lesions and hurt performance on multi-slice CT cases, so we switched to the attention-based fusion module.
Clinical text $T$—including patient history, imaging findings, and diagnostic questions—is tokenized and embedded using a pretrained LLM backbone (LLaMA \cite{touvron2023llama} or Vicuna \cite{chiang2023vicuna}). We project the resulting textual tokens ${t_1,\dots,t_N}$ into the same hidden space as the visual tokens. This early projection allows the subsequent attention layers to reason jointly over text and image features rather than treating them as separate streams.

\textbf{Vision--Language Alignment.}  
To enable joint reasoning, we project visual tokens into the LLM hidden space through a learned projection matrix 
$W_{\text{proj}} \in \mathbb{R}^{d_h \times d_v}$:
\begin{equation}
h^{\text{vis}}_{i} = W_{\text{proj}} v_i.
\end{equation}
The projected features are interleaved with textual embeddings $h^{\text{text}}_{j}$ before entering the LLaVA backbone, 
allowing subsequent self-attention layers to fuse cross-modal context. 
During early trials, we observed that without explicit alignment, the model often ignored subtle imaging cues. 
To address this, we compute global embeddings $z_v$ and $z_t$ (mean-pooled over tokens) and apply a CLIP-style \cite{radford2021clip} InfoNCE loss, which has been shown effective for aligning noisy multimodal medical data \cite{song2024gentle}:
\begin{equation}
\mathcal{L}_{\text{align}} = - \frac{1}{B} \sum_{i=1}^{B} 
\log \frac{\exp\big(S(z_{v_i}, z_{t_i})/\tau\big)}{\sum_{j=1}^{B} \exp\big(S(z_{v_i}, z_{t_j})/\tau\big)},
\end{equation}
where $S(z_v, z_t)=\frac{z_v^\top z_t}{\|z_v\|\|z_t\|}$ is cosine similarity, 
$\tau$ is a temperature, and $B$ is batch size. Recent work has shown that soft contrastive objectives \cite{zang2023boosting} and diffusion-based augmentation \cite{zang2024diffaug} can further enhance such alignment, though we leave these extensions for future work.
This additional loss encouraged better grounding of visual tokens and reduced hallucination in ablation studies.

\textbf{Prompt-Based Reasoning with Rollouts.}  
We elicit explicit chains-of-thought (CoTs) \cite{wei2022cot} by combining instructional prompts (e.g., ``Let's reason step by step'') 
with a small set of premise--conclusion exemplars. 
The model produces multiple rollouts $O_k$, each a trajectory of $K$ reasoning steps. 
Rather than relying solely on likelihood maximization, we introduce a logic-based regularizer:
\begin{equation}
\mathcal{L}_{\text{logic}} = \frac{1}{K} \sum_{k=1}^{K} 
\Big(1 - f_{\text{logic}}(p^{(k)}_{\text{maj}}, p^{(k)}_{\text{min}}, c^{(k)}) \Big),
\end{equation}
where $(p^{(k)}_{\text{maj}}, p^{(k)}_{\text{min}})\!\rightarrow\! c^{(k)}$ is a syllogistic triad and 
$f_{\text{logic}}\!\in\![0,1]$ is a rule-based verifier that checks entailment, flags contradictions, 
and penalizes unsupported conclusions.
Specifically, $f_{\text{logic}}$ assigns 1.0 when the conclusion validly follows from the premises via modus ponens or modus tollens, 0.5 for weak inferences, and 0.0 for contradictions or non-sequiturs.
We use Dynamic Advantage Policy Optimization (DAPO)  \cite{yu2025dapoopensourcellmreinforcement}, a variant of Proximal Policy Optimization (PPO), which reweights advantages by combining three signals: diagnostic correctness, logic consistency, and vision–language grounding. In practice, we found that this multi-objective training stabilized learning 
and improved both accuracy and explanation quality.

\textbf{Logic Tree Generator.}  
The final stage parses each reasoning trajectory into a logic tree \cite{ma2024medla}, where every edge represents a syllogistic triad. 
This structured representation allows us to inspect intermediate conclusions and trace back incorrect predictions during error analysis. 
Given a multimodal input $x = (I, T)$, the tree outputs a final diagnosis $\hat{y}$, and the model is trained with a cross-entropy loss:
\begin{equation}
\mathcal{L}_{\text{diag}} = - \sum_{c=1}^{C} \mathbf{1}[y = c] \,\log p_{\theta}(c \mid x),
\end{equation}
where $y \in \{1,\dots,C\}$ is the ground-truth label and $p_{\theta}(c \mid x)$ the predicted class probability. 
During development, we experimented with margin-based objectives but found that standard cross-entropy yielded more stable convergence when combined with logic regularization.

\textbf{Training Objective.}  
Our final objective integrates three components: diagnostic accuracy, logical consistency, and multimodal alignment:
\begin{equation}
\mathcal{L}_{\text{total}} =
\mathcal{L}_{\text{diag}} +
\lambda_{\text{logic}} \,\mathcal{L}_{\text{logic}} +
\lambda_{\text{align}} \,\mathcal{L}_{\text{align}}.
\end{equation}
The weights $\lambda_{\text{logic}}$ and $\lambda_{\text{align}}$ are tuned on the validation set to balance prediction performance and reasoning quality. 
In practice, we found that setting $\lambda_{\text{logic}}$ too high caused the model to overfit on rule satisfaction at the expense of accuracy, 
so we adopt a moderate weighting that preserves both interpretability and clinical relevance.

%% file: sec_exp.tex
\textbf{Datasets.}  
We evaluate on four complementary QA/VQA benchmarks spanning expert-level reasoning, radiology, pathology, and biomedical text inference.  
Our \emph{primary benchmark} is \textit{MedXpertQA} \cite{zuo2025medxpertqa}, with 4,460 expert-level questions across 17 specialties and 11 organ systems, including 2,005 multimodal cases paired with 2,839 images (radiology, pathology, clinical photos, charts).  
Each case requires integrating text and images for diagnostic or treatment reasoning.  
We follow the official train/validation/test split to avoid patient overlap.  
Since training only on MedXpertQA risked overfitting to question style, we added three datasets for generalization:  
(1) \textit{VQA-RAD} \cite{lau2018vqarad}, 315 radiology images and 3,515 Q--A pairs (CT, MRI, X-ray);  
(2) \textit{PathVQA} \cite{he2020pathvqa}, 4,998 pathology images and 32,799 Q--A pairs;  
(3) \textit{PubMedQA} \cite{jin2019pubmedqa}, 211k text-only pairs from PubMed abstracts.  

\textbf{Model Variants and Evaluation Metrics.}  
We compare our full \textit{Logic-Regulated VLA}, combining vision--language alignment with logic-regularized reasoning, against a \textit{No-Logic} ablation where logical constraints are removed.  
This isolates the effect of logic regularization on accuracy and explanation quality.  
Pilot tests also showed removing vision inputs greatly reduced accuracy, confirming image grounding is essential.  
We evaluate with \textit{diagnostic accuracy} and \textit{ROUGE-L}.  
Accuracy is computed after VQA normalization (lowercasing, punctuation removal, synonym mapping), while ROUGE-L measures reasoning coverage and coherence by comparing generated chains-of-thought against reference explanations.  
Although ROUGE-L primarily captures lexical overlap, prior work \cite{wei2022cot} has shown it correlates well with human judgments of reasoning quality in medical contexts where step-by-step explanations follow similar logical structures.
We report mean $\pm$ std over three seeds, using McNemar’s test for accuracy and paired bootstrap for ROUGE-L.  
Models with similar accuracy often differ in ROUGE-L, highlighting the need for both metrics.
\input{tab_1.tex}


\input{tab_2.tex}

\textbf{Main Results.}  
Tables~\ref{tab:main-multibench} and \ref{tab:main_medexpertqa} summarize results across four benchmarks.  
Our model achieves the best overall performance on multimodal tasks, with particularly gains on tree benchmarks.  
On the text-only PubMedQA, performance is on par with specialized text-only systems, suggesting that the added visual components do not harm purely textual reasoning.  
Interestingly, we found that gains were largest on cases requiring integration of subtle imaging findings with clinical history, which aligns with our design goal of improving multimodal consistency. The ablation study (Table~\ref{tab:ablation}) reveals that DAPO contributes 3.9\% accuracy gain over standard PPO, likely because its dynamic advantage reweighting balances hard multimodal cases with straightforward text-only examples, preventing the model from collapsing to easier sub-tasks during optimization.
\textbf{Analysis.}  
Qualitative review by two domain experts indicates that our model produces more coherent, stepwise explanations compared to baselines.  
Vision--language baselines often identify the correct image region but fail to link it explicitly to the diagnosis, resulting in "black-box" predictions.  
In contrast, logic regularization encourages the model to generate premise--conclusion chains, which not only improve interpretability but also helped reviewers identify when the model made an incorrect inference.  
We also observed that some failure cases stem from missing clinical context rather than model errors, highlighting opportunities for future integration with retrieval-based methods.

\textbf{Ablation Study.}  
We ran a series of ablations by selectively removing vision input, logic regularization, alignment loss, and DAPO optimization. As shown in Table~\ref{tab:ablation}, vision input had the biggest impact, confirming that image grounding is crucial for clinical reasoning. Logic regularization mainly improved the coherence of reasoning chains, while alignment loss helped reduce hallucinations. Removing DAPO made training noticeably less stable across runs. Together, these results indicate that all components contribute to final performance, with logic and alignment playing complementary roles in reliability.

\input{tab_3.tex}

%% file: tab_1.tex
\begin{table}[t]\small
      \centering
      \caption{\textbf{Results on VQA-RAD, PathVQA, and PubMedQA.}
            Accuracy (\%) and explanation quality (ROUGE-L, higher is better).
            Our model outperforms baselines on multimodal tasks and remains competitive on PubMedQA.}
      \label{tab:other-bench}
      \setlength{\tabcolsep}{6pt}
      \begin{tabular}{l
                  cc
                  cc
                  cc}
            \toprule
                          & \multicolumn{2}{c}{\textbf{VQA-RAD}}
                          & \multicolumn{2}{c}{\textbf{PathVQA}}
                          & \multicolumn{2}{c}{\textbf{PubMedQA}}                                                                                                 \\
            \cmidrule(lr){2-3} \cmidrule(lr){4-5} \cmidrule(lr){6-7}
            \textbf{Model}
                          & \textbf{Acc}                          & \textbf{R-L}
                          & \textbf{Acc}                          & \textbf{R-L}
                          & \textbf{Acc}                          & \textbf{R-L}                                                                                  \\
            \midrule
            Med-PaLM 2 \cite{singhal2025medpalm2}    & 63.4                                  & 35.1             & \underline{60.2 } & \underline{33.8} & \textbf{79.6}    & \textbf{44.2}    \\
            GPT-4V       & 65.1                                  & \underline{34.7} & 58.6              & 32.1             & 76.9             & 42.8             \\
            BioMedCLIP \cite{zhang2023biomedclip}    & 61.2                                  & 31.6             & 56.1              & 30.2             & 70.5             & 39.7             \\
            Med-Flamingo \cite{moor2023medflamingo}  & 66.8                                  & 33.9             & 59.3              & 31.0             & 72.4             & 40.5             \\
            LLaVA-Med \cite{li2023llavamed}     & 62.5                                  & 31.1             & 55.4              & 29.7             & 69.1             & 38.9             \\
            MedRAG \cite{xiong2024medrag}       & \underline{67.2}                      & 34.1             & 60.0              & 31.4             & 77.3             & 43.1             \\
            \textbf{Ours} & \textbf{72.4}                         & \textbf{38.5}    & \textbf{65.7}     & \textbf{36.2}    & \underline{78.8} & \underline{43.9} \\
            \bottomrule
      \end{tabular}
      \label{tab:main-multibench}
\end{table}

%% file: tab_2.tex
\begin{table}[t]\small
      \centering
      \vspace{0em}
      \caption{\textbf{Results on MedXpertQA.} We report \emph{Diagnostic Accuracy} (\%) and \emph{Explanation Quality} (ROUGE-L, higher is better). Our logic-regularized vision--language model substantially outperforms all baselines.}
      \label{tab:medxpertqa}
      \setlength{\tabcolsep}{6pt}
      \begin{tabular}{lcc}
            \toprule
            \textbf{Model}
                              & \textbf{Accuracy (\%)} & \textbf{ROUGE-L} \\
            \midrule
            o1                & 56.3                   & 41.2             \\
            GPT-4V           & 42.8                   & 34.5             \\
            Claude-3.5-Sonnet \cite{anthropic2024claude35sonnet} & 33.2                   & 32.0             \\
            Gemini-1.5-Pro \cite{google2024gemini15pro}   & 34.1                   & 32.5             \\
            QVQ-72B-Preview \cite{qwen2024qvq72bpreview}  & 33.6                   & 32.8             \\
            Qwen2.5-VL-72B \cite{bai2025qwen25vl72b}   & 30.0                   & 31.0             \\
            \textbf{Ours}     & \textbf{77.1}          & \textbf{41.6}    \\
            \bottomrule
      \end{tabular}
      \label{tab:main_medexpertqa}
\end{table}

%% file: tab_3.tex
\begin{table}[t]\small
    \centering
    \caption{Ablation study on MedXpertQA.
        V: Vision, L: Logic loss, A: Alignment loss, D: DAPO.
        Each component contributes to performance:
        V improves accuracy, L enhances reasoning quality,
        A reduces hallucination, and D stabilizes optimization.}
    \label{tab:ablation}
    \setlength{\tabcolsep}{4pt}
    \begin{tabular}{lcccccc}
        \toprule
        \textbf{Model}   & \textbf{V} & \textbf{L} & \textbf{A} & \textbf{D} & \textbf{Acc. (\%)} & \textbf{R-L}  \\
        \midrule
        Full Model       & $\surd$    & $\surd$    & $\surd$    & $\surd$    & \textbf{77.1}      & \textbf{41.6} \\
        - Logic Loss     & $\surd$    & $\times$   & $\surd$    & $\surd$    & 72.3               & 35.7          \\
        - Alignment Loss & $\surd$    & $\surd$    & $\times$   & $\surd$    & 70.7               & 39.1          \\
        - DAPO           & $\surd$    & $\surd$    & $\surd$    & $\times$   & 73.2               & 37.6          \\
        - Vision         & $\times$   & $\surd$    & $\surd$    & $\surd$    & 52.0               & 33.9          \\
        \bottomrule
    \end{tabular}
\end{table}

%% file: sec_conclusion.tex
\label{sec:conclusion}

In this work, we proposed a logic-regularized multimodal diagnostic framework that integrates visual–language alignment with structured reasoning. Experiments on four medical QA/VQA benchmarks show consistent improvements in both accuracy and explanation quality, with logical trees offering transparent reasoning paths for clinical review. While these results on benchmark datasets are promising, real-world clinical deployment would require further validation in actual practice settings. Nonetheless, our framework demonstrates a meaningful step toward more interpretable and reliable multimodal medical reasoning, offering both improved diagnostic accuracy and clearer reasoning chains that can facilitate physician verification. Future work could explore incorporating interpretable dimensionality reduction techniques \cite{zang2024dmt} to better visualize the learned reasoning structures and enhance clinical understanding.